# Redesigning Decision Matrix Method with an indeterminacy-based inference process

Jose L. Salmeron[a*] and Florentin Smarandache[b]

[a] Pablo de Olavide University at Seville (Spain)

[b] University of New Mexico, Gallup (USA)

ABSTRACT

For academics and practitioners concerned with computers, business and mathematics, one central issue is supporting decision makers. In this paper, we propose a generalization of Decision Matrix Method (DMM), using Neutrosophic logic. It emerges as an alternative to the existing logics and it represents a mathematical model of uncertainty and indeterminacy. This paper proposes the Neutrosophic Decision Matrix Method as a more realistic tool for decision making. In addition, a de-neutrosophication process is included.

Mathematics Subject Classification: Neutrosophic Logic.

---

\* Corresponding author
  e-mail address: salmeron@upo.es

# 1. INTRODUCTION

For academics and practitioners concerned with computers, business and mathematics, one central issue is supporting decision makers. In that sense, making coherent decisions requires knowledge about the current or future state of the world and the path to formulating a fit response (Zack, 2007).

The authors propose a generalization of Decision Matrix Method (DMM), or Pugh Method as sometimes is called, using Neutrosophic logic (Smarandache, 1999). The main strengths of this paper are two-folds: it provides a more realistic method that supports group decisions with several alternatives and it presents a de-neutrosophication process. We think this is an useful endeavour.

The remainder of this paper is structured as follows: Section 3 reviews Decision Matrix Method; Section 3 shows a brief overview of Neutrosophic Logic and proposes Neutrosophic Decision Matrix Method and de-neutrosophication process; the final section shows the paper's conclusions.

# 2. DECISION MATRIX METHOD BACKGROUND

Decision Matrix Method (DMM) was developed by Stuart Pugh (1996) as an approach for selecting concept alternatives. DMM is a method (Murphy, 1979) that allows decision makers to systematically identify and analyze the strength of relationships between sets of information. This technique is especially interesting for looking at large numbers of factors and assessing each relative importance. Furthermore, DMM is a method for alternative selection using a scoring matrix. DMM is often used throughout planning activities to select product/service features and goals and to develop process stages and weight options.

DMM is briefly exposed. At the first time an evaluation team is established. Firstly, the team selects a list of weighted criteria and then evaluates each alternative against the previous criteria. That election could be done using any technique or mix of them (discussion meetings, brainstorming, and so on). This one must be refined in an iterative process.

The next step is to assign a relative weight to each criterion. Usually, ten points are distributed among the criteria. This assignment must be done by team consensus. In addition, each team member can assign weights by himself, then the numbers for each criterion are added for a composite criterion weighting.

Follow that, L-shaped matrix is drawn. This kind of matrix relates two groups of items to each other (or one group to itself). In the last step, the alternatives are scored relative to criteria.

Figure 1. Building a Decision Matrix

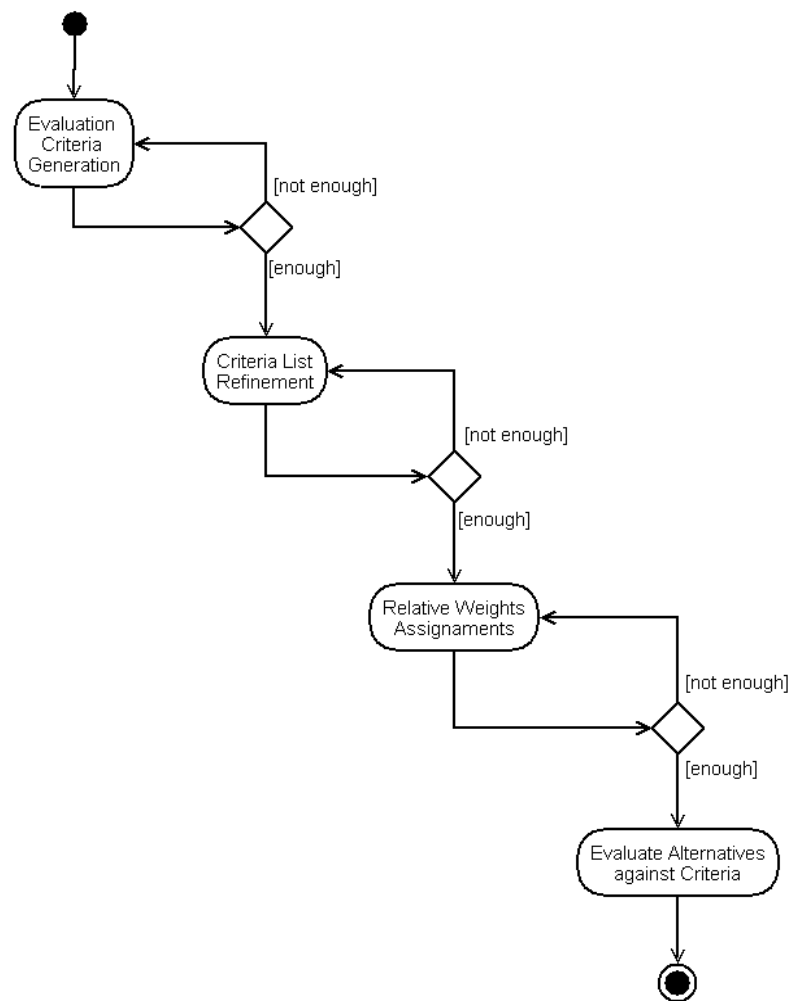

Some options are showed in Table 1.

Table 1. Assessing alternatives

| | Method | Values range |
|---|---|---|
| 1 | Rating scale for each alternative. | For example {1=low, 2=medium, 3=high} |
| 2 | For each criterion, rank-order all alternatives according to each fits the criterion. | Order them with 1 being the option that is least fit to criterion. |
| 3 | Establish a reference. It may be one of the alternatives or any current product/service. For each criterion, rate each other alternative in comparison to the baseline. | For example: Scores of {-1=worse, 0=same, +1=better} Wider scales could be used. |

At the end, multiply each alternative's rating by its weight. Add the points for each alternative. The alternative with the highest score will be the team's proposal.

Let C be the criteria vector of a DMM. $C = (c_1, c_2, ..., c_n)$ where $c_j$ belongs to the criteria dominion of the problem and n is the total number of criteria.

Let W be the weights criteria vector of a DMM. $W = (w_1, w_2, ..., w_n)$ where $w_j \in [0, N) | N \neq \infty$.

Let $A_i$ be the rating vector of i alternative. $A_i = (a_1, a_2, ..., a_n)$ where $a_m \in \{-1, 0, 1\}$.

Consider the matrix D be defined by $D = (a_{ij})$ where $a_{ij}$ is the rating of alternative i to the criterion j, $a_{ij} \in \{-1, 0, 1\}$. D is called the rating matrix of the DMM.

Consider the vector S be defined by $S = W \times D$, being $D = (s_1, s_2, ..., s_m)$ where $s_k$ is the product of weight i by alternative j and m is the number of alternatives.

$$\begin{pmatrix} s_1 & s_2 & ... & s_m \end{pmatrix} = \begin{pmatrix} w_1 & w_2 & ... & w_n \end{pmatrix} \times \begin{pmatrix} a_{11} & ... & a_{m1} \\ a_{12} & ... & a_{m2} \\ ... & ... & ... \\ a_{1n} & ... & a_{mn} \end{pmatrix}$$

The highest $s_k$ will be the team's proposal for the problem analyzed. Additionally, alternatives have been ranked by the team.

It is important to note that $s_k$ measures only rate of alternative j respect to weight i, till now any scholar has not contemplated the indeterminacy of any relation between alternatives and criteria.

When we deal with unsupervised data, there are situations when team can not to determine any rate. Our proposal includes indeterminacy in DMM generating more realistic results. In our opinion, including indeterminacy in DMM is an useful endeavour.

## 3. NEUTROSOPHIC LOGIC FUNDAMENTALS

Neutrosophic Logic (Smarandache, 1999) emerges as an alternative to the existing logics and it represents a mathematical model of uncertainty, and indeterminacy. A logic in which each proposition is estimated to have the percentage of truth in a subset T, the percentage of indeterminacy in a subset I, and the percentage of falsity in a subset F, is called Neutrosophic Logic. It uses a subset of truth (or indeterminacy, or falsity), instead of using a number, because in many cases, humans are not able to exactly determine the percentages of truth and of falsity but to approximate them: for example a proposition is between 30-40% true.

The subsets are not necessarily intervals, but any sets (discrete, continuous, open or closed or half-open/ half-closed interval, intersections or unions of the previous sets, etc.) in accordance with the given proposition. A subset may have one element only in special cases of this logic. It is imperative to mention here that the Neutrosophic logic is a strait generalization of the theory of Intuitionist Fuzzy Logic.

According to Ashbacher (2002), Neutrosophic Logic is an extension of Fuzzy Logic (Zadeh, 1965) in which indeterminacy is included. It has become very essential that the notion of neutrosophic logic play a vital role in several of the real world problems like law, medicine, industry, finance, IT, stocks and share, and so on. Static context of Neutrosophic logic is showed in Figure 2.

Figure 2. Static context of Neutrosophic logic

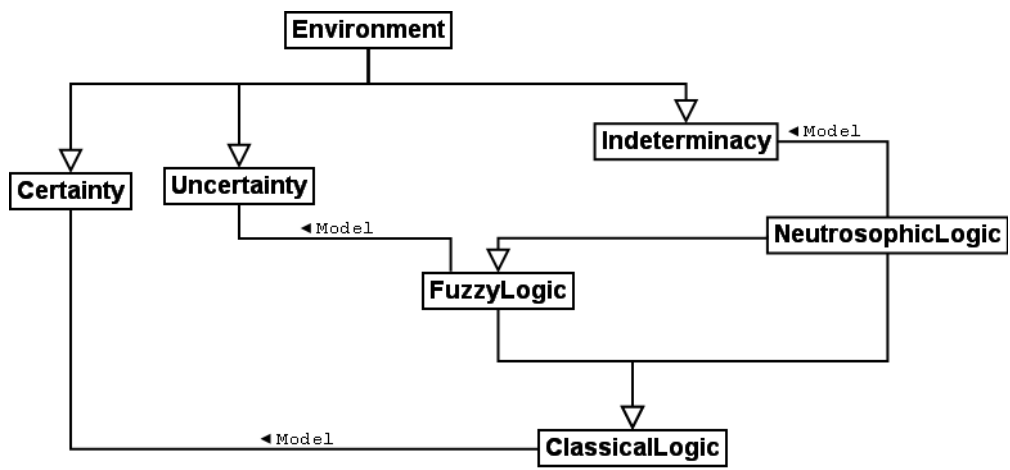

Fuzzy theory measures the grade of membership or the non-existence of a membership in the revolutionary way but fuzzy theory has failed to attribute the concept when the relations between notions or nodes or concepts in problems are indeterminate. In fact one can say the inclusion of the concept of indeterminate situation with fuzzy concepts will form the neutrosophic concepts. In NL each proposition is estimated to have the percentage of truth in a subset T, the percentage of indeterminacy in a subset I, and the percentage of falsity in a subset F.

We use a subset of truth (or indeterminacy, or falsity), instead of a number only, because in many cases we are not able to exactly determine the percentages of truth and of falsity but to approximate them: for example a proposition is between 30-40% true and between 60-70% false, even worst: between 30-40% or 45-50% true (according to various analyzers), and 60% or between 66-70% false. The subsets are not necessary intervals, but any sets (discrete, continuous, open or closed or half-open/half-closed interval, intersections or unions of the previous sets, etc.) in accordance with the given proposition.

A subset may have one element only in special cases of this logic. Statically T, I, F are subsets, but dynamically they are functions/operators depending on many known or unknown parameters.

Constants (T, I, F) truth-values, where T, I, F are standard or non-standard subsets of the non-standard interval $]^-0,1^+[$, where $n_{inf}$ = inf T + inf I + inf F ≥ $^-0$, and $n_{sup}$ = sup T + sup I + sup F ≤ $3^+$. Statically T, I, F are subsets, but dynamically T, I, F are functions/operators depending on many known or unknown parameters.

The NL is a formal frame trying to measure the truth, indeterminacy, and falsehood. The hypothesis is that no theory is exempted from paradoxes, because of the language imprecision, metaphoric expression, various levels or meta-levels of understanding/interpretation which might overlap.

### 3.1. Using indeterminacy in Decision Matrix Method

We propose a redesign of the DMM called Neutrosophic Decision Matrix Method (NDMM). This proposal includes indeterminacy in alternatives' rating and not is used to weights. It is because weights are the quantified value of criteria. They are selected by the team. Therefore, an indeterminacy weight has no sense. On the other hand, it is possible to consider indeterminacy to alternatives rating.

A Neutrosophic Decision Matrix is a neutrosophic matrix with neutrosophic values (alternatives ratings or indeterminacies as elements). Consider the matrix D be defined by $D = (a_{ij})$ where $a_{ij}$ is the neutrosophic value of alternative i to the criterion j. D is called the rating matrix of the NDMM. In that sense, $a_{ij} \subset [-1,1] \cup I$. We would interpret this expression as representing the total group of numbers as the union of two other groups. The first interval would start at -1 and proceed toward +1. The second would be an indeterminacy value.

The total set of numbers would be all those in the first group along with the indeterminacy value. Note that $I \in [-1,1]$, since it is an indeterminate value in that interval. In fact, we have that $a_{ij} = \{x | -1 \leq x \leq 1\}$.

In addition, we propose a de-neutrosophication process in NDMM. This one is based on max-min values of $I$. A neutrosophic value is transformed in an interval with two values, the maximum and the minimum value for $I$. In that sense, the neutrosophic scores will be an area, where the upper limit has $I = 1$ and the lower limit has $I = -1$. The solution set is $\chi = \bigcup_{j=1}^{n} s_j$, where j is the alternatives number and s is the score of each one. Any $s_k | k \neq j$ belongs to the complement of $\chi^c$. Alternative selected is the global maximum in $\chi$. It is an alternative $A_m$ where $s_m^* \geq s_i | \forall i ; i,m \in \chi$. De-neutrosophication process will be applied within the following application.

$s_m^*$ is a line (y axis value fixed) represented the score of alternative $A_m$. It is possible that $s_i \in s_j$, since $s_i$ is a line and $s_j$ is an area after de-neutrosophication process. We select according to

$$\begin{cases} s_i & if \quad s_i > \dfrac{\max s_j - \min s_j}{2} + k \, | \, k \leq (\max s_j - s_i) \\ s_j & otherwise \end{cases}$$

### 3.2. An application

This example illustrates the improvements of NDMM versus DMM. NDMM proposal allows to represent indeterminacy in a decisional framework. Let C be the criteria vector of a decision problem. $C = (c_1, c_2, c_3, c_4)$ where $c_j$ belongs to the criteria dominion of the problem.

Let W be the weights criteria vector of a DMM. $W = (w_1, w_2, w_3, w_4)$. We have used a three-valued scale from 1 (less importance) to 3 (more importance).

$$\begin{cases} w_1 = 3 \\ w_2 = 3 \\ w_3 = 2 \\ w_4 = 1 \end{cases}$$

Three different alternatives are been considering. We call it $A_i$, where i is the order of each one.

Consider the following Neutrosophic Decision Matrix where alternatives and ratings are showed. Each column represents the ratings for an alternative and each row gives the criterion ratings for all the alternatives.

$$Neutrosophic\ Decision\ Matrix = \begin{pmatrix} 5 & 6 & 7 \\ 2 & I & 5 \\ 10 & 4 & I \\ 3 & 2 & 7 \end{pmatrix}$$

We have used a scale from 1 (less fit) to 10 (more fit). Indeterminacy is introduced in the second alternative (second criterion) and the third alternative (third criterion).

We show the $S$ vector with the product of weight i by alternative j as a result.

$$(49 \quad 28+3I \quad 40+2I) = (3 \quad 3 \quad 2 \quad 1) \times \begin{pmatrix} 5 & 6 & 7 \\ 2 & I & 5 \\ 10 & 4 & I \\ 3 & 2 & 7 \end{pmatrix}$$

The neutrosophic score of each alternative is showed.

$$scores_n = \begin{cases} A_1 = 44 \\ A_2 = 28 + 3I \\ A_3 = 43 + 2I \end{cases}$$

If we consider scores got for second and third alternatives as equations the representation would be the showed in Figure 3. Obviously, $A_1$ is the best option.

Figure 3. Alternatives' neutrosophic scores

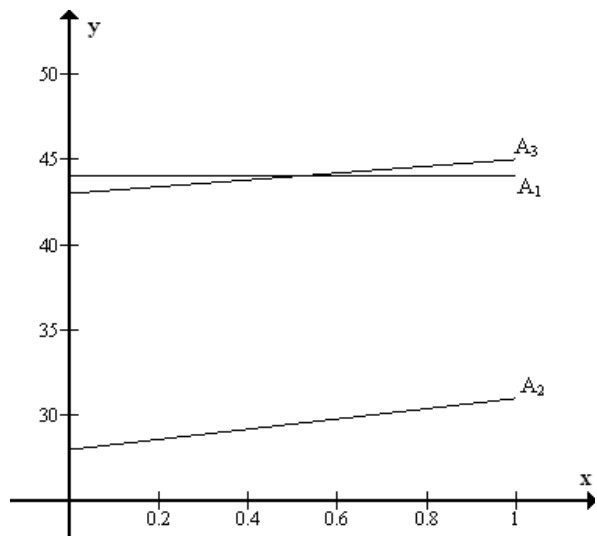

The next step is the de-neutrosophication process. We replace $I \in [0,1]$ both maximum and minimum values.

$$scores_d = \begin{cases} A_1 = 44 \\ A_2 = [28, 31] \\ A_3 = [43, 45] \end{cases}$$

Figure 4 shows the de-neutrosophic results. The results show alternatives 2 and 3 as areas. In this case $0 \leq k \leq 1$. $A_3$ will be selected if and only if $k > 0$. It is more realistic view from DMM.

Figure 4. Alternatives' de-neutrosophic scores

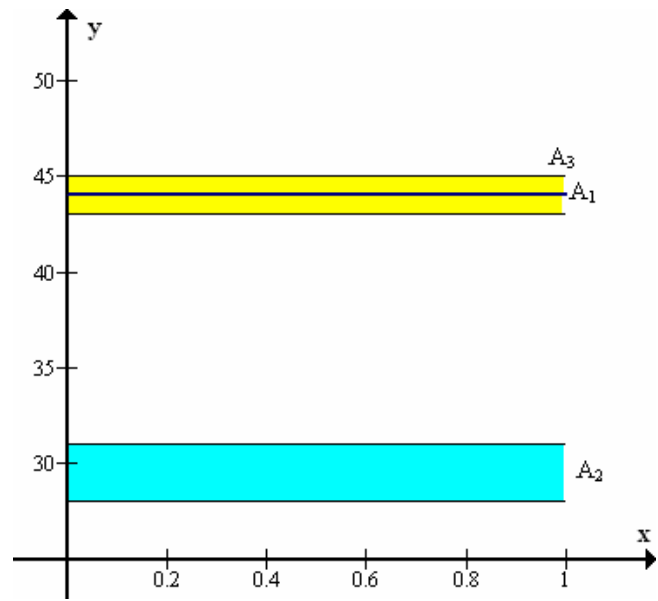

## 4. CONCLUSIONS

Numerous scientific publications address the issue of decision making in every fields. But, little efforts have been done for processing indeterminacy in this context. This paper shows a formal method for processing indeterminacy in Decision Matrix Method and include a de-neutrosophication process.

The main outputs of this paper are two-folds: it provides a neutrosophic tool for decision making and it also includes indeterminacy in a decision tool. In this paper a renewed Decision Matrix Method has been proposed. As a methodological support, we have used Neutrosophic Logic. This emerging logic extends the limits of information for supporting decision making and so on.

Using NDMM decision makers are not forced to select ratings when their knowledge is not enough for it. In that sense, NDMM is a more realistic tool since experts' judgements are focused on their expertise.

Anyway, more research is needed about Neutrosophic logic limit and applications. Incorporating the analysis of NDMM, the study proposes an innovative way for decision making.